\documentclass[runningheads]{llncs}
\usepackage{etoolbox}
\AtBeginEnvironment{thebibliography}{\footnotesize}
\usepackage[T1]{fontenc}
\usepackage{amsmath}
\usepackage{amssymb}
\usepackage{pifont} % for checkmark and xmark
\newcommand{\cmark}{\ding{51}}
\newcommand{\xmark}{\ding{55}}

\usepackage{orcidlink}
\usepackage{graphicx,verbatim}
\usepackage{booktabs}
\usepackage{makecell}
\usepackage{caption}

\begin{document}

\title{GraphSVR: q-Space--Aware Graph-Based Slice-to-Volume Registration for Diffusion MRI}
\titlerunning{GraphSVR: Slice-to-Volume Registration for Diffusion MRI}

\author{
Noga Kertes\inst{1,2}\orcidlink{0009-0000-2412-2019} \and
Daphna Link Sourani \inst{1,2}\orcidlink{0000-0003-2422-6795} \and
Alex M. Bronstein\inst{3,4}\orcidlink{0000-0001-9699-8730} \and
Moti Freiman\inst{1,2}\orcidlink{0000-0003-1083-1548}
}

\institute{
Faculty of Biomedical Engineering, Technion -- Israel Institute of Technology, Haifa, Israel
\and
The May-Blum-Dahl MRI Research Center, Faculty of Biomedical Engineering, Technion -- Israel Institute of Technology, Haifa, Israel
\and
The Taub Faculty of Computer Science, Technion -- Israel Institute of Technology, Haifa, Israel
\and
Institute of Science and Technology Austria (ISTA), Klosterneuburg, Austria\\
\email{noga.kertes@campus.technion.ac.il}
}

\authorrunning{N. Kertes et al.}

\maketitle

\begin{abstract}
Diffusion-weighted imaging (DWI) remains highly vulnerable to subject motion, particularly in time-efficient protocols and in motion-prone populations. While slice-to-volume registration (SVR) can mitigate inter-slice and inter-stack misalignment, diffusion MRI introduces additional complexity due to diffusion-direction–dependent contrast and the requirement to align dozens of measurements within a common reference frame, effectively yielding a 4D registration problem. Existing approaches rely primarily on sequential modeling or pairwise similarity and often degrade under sparse gradient sampling or severe motion.
We introduce GraphSVR, a q-space–aware graph-based framework for 4D SVR registration in DWI. GraphSVR represents slice groups as nodes in an acquisition-structured graph, with edges encoding temporal proximity, spatial slice geometry and diffusion encoding relationships. A graph neural network predicts globally consistent stack-wise rigid motion, optimized in a self-supervised, zero-shot manner using only an anatomical reference image, without requiring paired ground-truth motion.
We evaluate GraphSVR using both fully synthetic diffusion simulations and realistic recombination-based simulations from real acquisitions with controllable motion severity and gradient sparsity. Performance is quantified using grid error (mm) and rotation error relative to known ground-truth transforms. Under severe motion, GraphSVR reduces grid error and rotation error by 73\% compared to FSL eddy, the standard DWI motion-correction method, with the largest gains observed in sparse-direction regimes. These results demonstrate that explicitly modeling acquisition structure through graph-based reasoning improves robustness and global consistency in 4D DWI motion estimation.Code is available at \url{https://github.com/nogakertes/GraphSVR.git}.

\end{abstract}

\section{Introduction}

Diffusion-weighted imaging (DWI) enables in vivo characterization of white-matter microstructure and connectivity \cite{le2001diffusion,johansen2013diffusion,mori2002imaging}. Accurate diffusion modeling requires tens of diffusion-weighted volumes acquired over extended scan times, making DWI highly susceptible to subject motion \cite{mukherjee2008diffusion,havsteen2017movement}. This limitation is particularly problematic in motion-prone populations, including pediatric and clinical cohorts, where motion degrades image quality, increases exclusion rates, and may require sedation \cite{gallo2023pediatric,artunduaga2021safety}. Although echo-planar imaging (EPI) enables rapid slice acquisition, inter-stack misalignment remains unavoidable \cite{stehling1991echo}. As a result, retrospective slice-to-volume registration (SVR) is commonly used to align 2D slice groups within a consistent 3D reference frame \cite{marami2016motion,8847637,xu2022svort,andersson2017towards,deprez2019higher}.

SVR in DWI is substantially more challenging than in structural MRI. Diffusion-direction–dependent contrast, varying b-values, and acquisition-order effects cause appearance changes across stacks \cite{andersson2017towards}. Motion must therefore be estimated consistently across multiple diffusion encodings and aligned to a shared anatomical reference, resulting in a 4D registration problem spanning space and q-space. Robustly coupling motion estimates under sparse gradient sampling and severe motion remains an open problem.

Existing methods address this setting only partially. Model-based q-space representations such as 3D-SHORE can synthesize diffusion-weighted contrasts, but their coefficients are estimated independently at each voxel and therefore do not directly enforce spatial consistency of slice-wise motion \cite{ozarslan2009threeDshore}. Eddymotion, now continued as NiFreeze, uses leave-one-volume-out signal prediction followed by volume-to-volume registration, and consequently does not explicitly estimate motion occurring between slice groups within a volume \cite{pisner2026nifreeze}. SVoRT \cite{xu2022svort} jointly estimates slice poses and reconstructs a latent 3D volume using a Transformer, but it targets a small number of orthogonal stacks and does not readily extend to DWI acquisitions with many similarly oriented, contrast-varying volumes. In DWI, FSL eddy \cite{andersson2017towards} is the standard for joint motion and eddy-current correction, leveraging cross-volume information; however, its performance degrades under sparse sampling  \cite{tax2022s} and severe motion. Sequential tracking approaches \cite{marami2016motion} improve robustness in sparse regimes but do not explicitly model global relationships across stacks and diffusion encodings.

We argue that DWI motion estimation is inherently relational: slice groups are coupled through acquisition timing, spatial geometry, and diffusion encoding. This structure is naturally represented as a graph. Graph neural networks (GNNs) aggregate node and edge information via message passing \cite{velivckovic2018graph}, enabling globally consistent, permutation-invariant estimation. Similar to pose-graph optimization in multi-view geometry \cite{li2021pogo}, DWI stacks can be viewed as multiple encoded views of the same anatomy, suggesting that graph-based reasoning is well suited to 4D SVR. Prior GNN-based registration methods \cite{yang2022graformerdir} focus on dense alignment of a few 3D volumes and do not address slice-level, time-ordered, contrast-varying DWI acquisitions.

We therefore introduce \textbf{GraphSVR}, a q-space–aware graph-based framework for 4D slice-to-volume registration in DWI. GraphSVR represents slice groups as nodes in an acquisition graph, with edges encoding temporal proximity, spatial adjacency, diffusion direction, and b-value similarity. A GNN predicts globally consistent stack-wise rigid motion in a self-supervised, zero-shot manner using only an anatomical reference image. This formulation explicitly couples spatial and q-space relationships to enforce global consistency under sparse sampling and severe motion.

We evaluate GraphSVR using recombination-based simulations from measured data and fully synthetic diffusion simulations with controlled motion severity and gradient sparsity. Under severe motion, GraphSVR reduces grid and rotation errors by up to \textbf{73\%} relative to FSL \textit{eddy}, with the largest gains in sparse-direction regimes. These results demonstrate that modeling acquisition structure through graph-based reasoning substantially improves robustness in 4D DWI motion estimation.

\section{Methods}
\begin{figure}[t]
    \centering
    \includegraphics[width=0.9\linewidth]{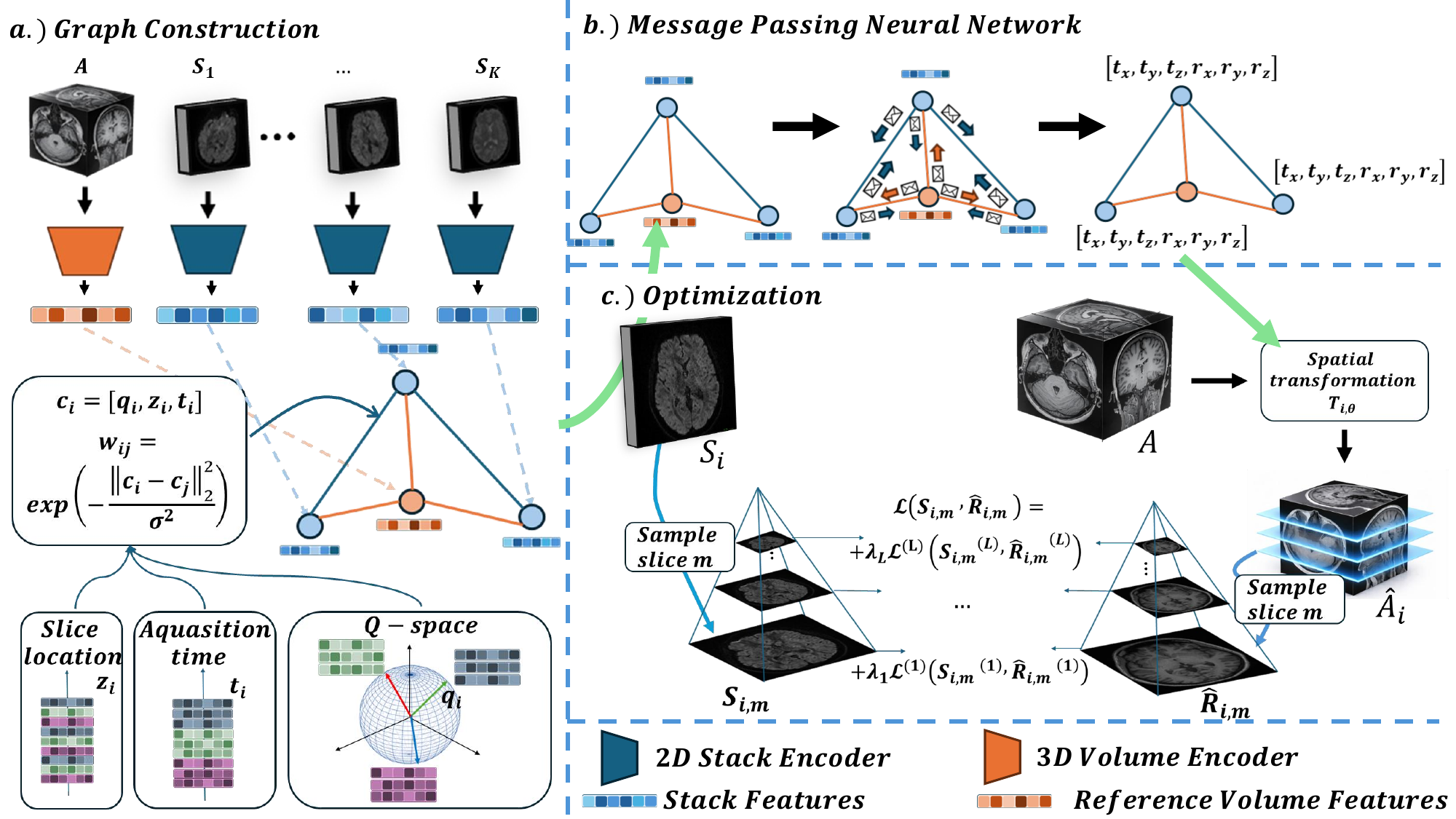}
    \caption{\textbf{Overview of GraphSVR.} Each slice group (stack) is a graph node; edges encode proximity in time, slice location, and q-space. A GNN predicts stack-wise rigid motion used to warp an anatomical reference into each stack. A multi-scale slice-wise similarity objective plus motion regularization enables subject-specific (zero-shot) optimization.}
    \label{fig:overview}
\end{figure}

\noindent\textbf{Acquisition and problem setup.}
A DWI acquisition consists of $K$ diffusion-weighted measurements with encodings $\{q_i\}_{i=1}^{K}$ ($b_i>0$), together with a small number of non-diffusion ($b=0$) images. With multiband factor $p$, each readout produces a slice group (stack) $S_i$ comprising $p$ interleaved slices acquired at known spatial locations under encoding $q_i$. We represent this acquisition as:
\[
\mathcal{S}=\{(I_i, t_i, z_i, q_i)\}_{i=1}^{K},
\]
where $I_i$ denotes the observed image data, $t_i$ the acquisition time, $z_i$ the physical slice locations, and $q_i$ the diffusion encoding. Given an anatomical reference volume $A$ (T1- or T2-weighted), GraphSVR estimates a rigid transform $T_i \in SE(3)$ for each stack. The transforms are obtained by optimizing a self-supervised objective that warps $A$ into the coordinate frame of each stack and maximizes image similarity.

\subsection{Graph construction}

Fig.~\ref{fig:overview}a presents the acquisition graph construction. We model the acquisition as a weighted graph $G=(V,E)$, where each node represents a stack and carries a feature vector $\mathbf{v}_i$ (Sec.~\ref{sec:architecture}). Edges encode pairwise relationships with non-negative weights $w_{ij}$.

We construct a $k$-nearest neighbor graph in a joint acquisition space. For each stack $i$, we define $\mathbf{c}_i=[\,\tilde{\mathbf{q}}_i,\tilde{\mathbf{z}}_i,\tilde{t}_i\,]$, where $\mathbf{q}_i\in\mathbb{R}^3$ denotes the diffusion gradient direction, $\mathbf{z}_i\in\mathbb{R}^{s}$ contains the physical slice positions of the $s$ slices in the stack, and $t_i$ denotes acquisition time in seconds. We z-normalize each component across stacks and compute an undirected $k$NN graph using Euclidean distance in this space. We define edge weights using an RBF kernel, with $\sigma^2$ set to the median squared pairwise distance across graph edges.

To provide a global anchor, we introduce a reference node $r$ corresponding to the anatomical volume and connect it to all stack nodes with unit weight. We add self-loops and apply symmetric GCN normalization, $\hat{w}_{ij}=w_{ij}/\sqrt{d_i d_j}$, where $d_i=\sum_j w_{ij}$.

\subsection{Network architecture}
\label{sec:architecture}

GraphSVR combines convolutional encoders with an attention-based graph network to predict rigid motion. For each stack $S_i$, we apply a lightweight 2D ResNet-style CNN \cite{he2016deep} independently to each slice and sum the resulting embeddings to obtain a stack-level descriptor $\mathbf{v}_i \in \mathbb{R}^{F}$. 

We encode the anatomical reference volume $A$ using a 3D CNN to produce a global embedding $\mathbf{v}_r \in \mathbb{R}^{F}$, which serves as the reference node feature.

We apply a multi-layer attention-based message passing network \cite{velivckovic2018graph,shi2020masked} that incorporates scalar edge weights during aggregation. A final MLP head regresses 6-DoF rigid motion parameters $\hat{\boldsymbol{\xi}}_i = [\hat{\mathbf{r}}_i, \hat{\mathbf{t}}_i] \in \mathbb{R}^{6}$ for each stack node, which we convert to $\hat{T}_i \in SE(3)$.

\subsection{Zero-shot self-supervised optimization}

We optimize GraphSVR per case without motion supervision. For each stack $S_i$, we estimate $T_i$ that maps the anatomical reference into the stack coordinate frame. We warp the reference volume as $\hat{A}_i=\mathrm{warp}(A;T_i)$, sample slices at the known locations as $\hat{R}_i=\mathrm{sample}(\hat{A}_i;z_i)$, and compute a slice-wise loss over the $s$ slices in the stack (Fig.~\ref{fig:overview}c).

Let $m\in\{0,\ldots,p-1\}$ denote the slice index within a stack of $p$ interleaved slices. The per-slice loss is defined as a weighted combination of local mutual information (LMI) \cite{chen2022transmorph} and normalized gradient fields (NGF) \cite{haber2006intensity}: $\mathcal{L}(S_{i,m},\hat{R}_{i,m})=\alpha\,\mathcal{L}_{\mathrm{LMI}}(S_{i,m},\hat{R}_{i,m})+\beta\,\mathcal{L}_{\mathrm{NGF}}(S_{i,m},\hat{R}_{i,m})$.

To avoid local minima, we optimize the loss using a multi-scale approach with $L$ scales:
\begin{equation}
\mathcal{L}_{\mathrm{ms}}(S_{i,m}, \hat{R}_{i,m})
=
\sum_{\ell=1}^{L}
\lambda_{\ell}
\Big(
\alpha\,\mathcal{L}_{\mathrm{LMI}}(S_{i,m}^{(\ell)}, \hat{R}_{i,m}^{(\ell)})
+
\beta\,\mathcal{L}_{\mathrm{NGF}}(S_{i,m}^{(\ell)}, \hat{R}_{i,m}^{(\ell)})
\Big).
\end{equation}

\section{Experiments}

\subsection{Data and motion simulation}

Quantitative evaluation of SVR in DWI is challenging because real scans lack ground-truth intra-acquisition motion. We therefore evaluate under controlled simulations with known transforms using two complementary strategies: (i) real-image recombination from measured data and (ii) fully synthetic data with controllable motion.

We acquired single-shot EPI DWI data from seven healthy volunteers on a Siemens Prisma 3T scanner at the May-Blum-Dahl MRI Research Center. For real-image simulations, one adult subject underwent five consecutive scans within a single session, changing head pose between scans while remaining still during each scan. Each scan included a T1-weighted anatomical reference (matrix $192{\times}256{\times}256$, voxel size $\sim$0.9–0.94 mm) and a DWI acquisition with 3 $b{=}0$ and 64 diffusion-weighted volumes at $b{=}1000$ s/mm$^2$ (matrix $128{\times}128{\times}82$, voxel size 1.76$\times$1.76$\times$1.6 mm). We estimated inter-scan rigid transforms by registering the T1 volumes using FreeSurfer \cite{reuter2010highly}. These registration-derived transforms serve as approximate ground truth, as they remain subject to residual volume-to-volume T1 registration error.

\emph{Real-image recombination:} We simulated intra-volume motion by assembling DWI acquisitions from slice groups drawn across different scans of the same subject. Because each scan was internally motion-free but differed in pose, this procedure preserves realistic DWI appearance while providing known stack-wise rigid transforms. 

\emph{Fully synthetic data:} Starting from the motion-free first scan, we fit a DTI model \cite{basser1994mr} and imposed smooth, time-varying rigid-motion trajectories across slice groups. We reoriented the tensor field using log-Euclidean interpolation \cite{arsigny2006log}, rotated diffusion encodings accordingly, and synthesized diffusion-weighted images via the DTI forward model. We added Rician noise (SNR=30) and evaluated sparse-direction regimes by subsampling gradient directions using approximately uniform subsets on the unit sphere \cite{jones1999optimal}. 

All simulations used multiband factor $p{=}4$ and included two $b{=}0$ volumes. For real-image recombination, we generated acquisitions with 12, 24, and 32 diffusion directions. For synthetic data, we evaluated the same direction counts at three motion-severity levels. In all experiments, the first time point was motion-free and aligned to the reference frame. In total, 21 real-image recombination simulations were generated (3 per subject), and 42 fully synthetic simulations were generated (6 per subject).

\subsection{Experimental Setup and Evaluation Protocol}

We optimized GraphSVR per case in a zero-shot manner with random initialization. We used Adam for up to 400 iterations (batch size 120, initial learning rate $10^{-3}$), together with a ReduceLROnPlateau scheduler (factor 0.5, patience 10) and early stopping after 15 epochs without improvement.

We set similarity weights to $\alpha=\beta=1$, constructed the acquisition graph with $k=4$ nearest neighbors, and used a GNN with two attention layers (four heads, hidden dimension 64). The multi-scale objective employed exponentially decaying pyramid weights $\lambda_\ell = 2^{-(\ell-1)}$. We selected hyperparameters, including $k$ and the number of pyramid levels, via empirical tuning on a held-out subset.

We implemented all experiments in PyTorch and PyTorch Geometric and executed them on a single NVIDIA L40 GPU (46\,GB VRAM).

We compared GraphSVR with FSL \textit{eddy} \cite{andersson2017towards}. To reduce confounding from distortion modeling, we used the linear eddy-current field model, disabled susceptibility correction, and re-referenced all \textit{eddy} motion estimates to the first time point (identity) to align reference frames.

In simulation experiments, we evaluated performance using \emph{grid error} (GE, mm) and \emph{rotation error} (RE, radians) relative to ground-truth transforms. 
Grid error was computed by applying the predicted and ground-truth affine transformations to a regular 3D grid spanning the image domain and averaging the Euclidean distance between corresponding transformed points. GE is reported in millimeters.
For each case and motion condition, we summarized performance by the median error across stacks and assessed significance using a paired two-sided Wilcoxon signed-rank test across cases.

\section{Results}
\subsection{Real-image recombination}
\begin{figure}[t]
    \centering
    \includegraphics[width=0.9\linewidth]{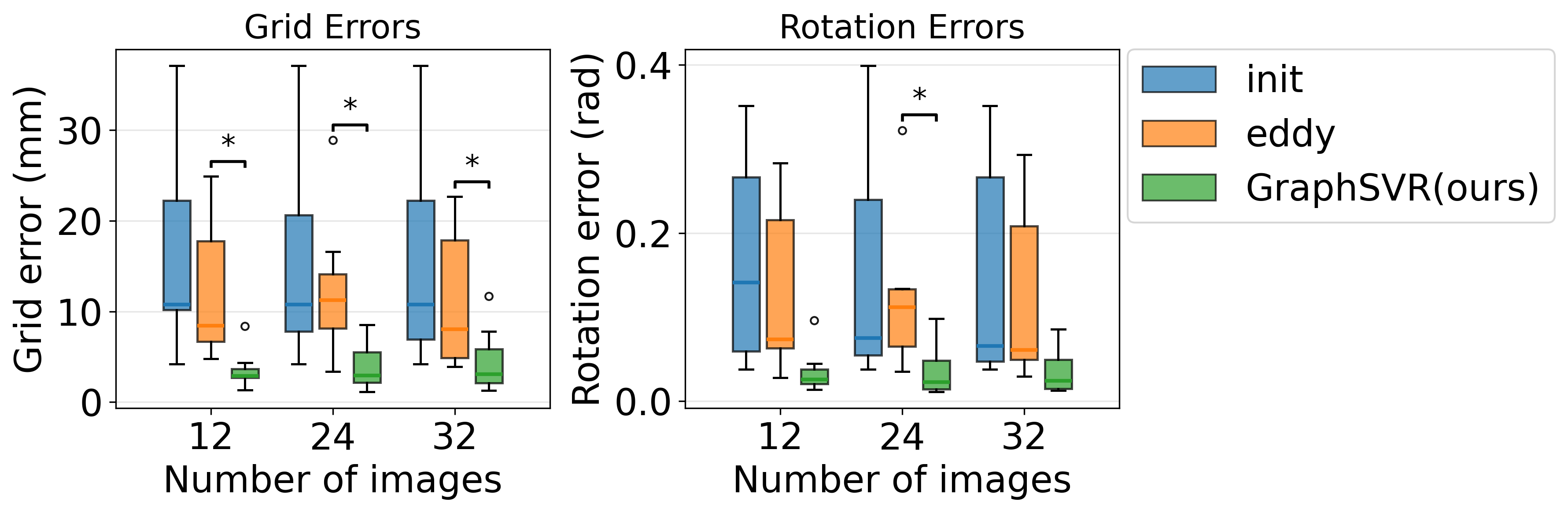}
    \caption{\textbf{Real-image recombination: translation and rotation errors.}
Grid error (GE, mm; left) and rotation error (RE, rad; right) for \texttt{init}, \textit{eddy}, and GraphSVR across acquisitions with 12, 24, and 32 diffusion directions. Boxplots show per-case median errors across stacks. Asterisks indicate settings where GraphSVR significantly outperforms \textit{eddy}.}
    \label{fig:real_sim_half_gridrot}
\end{figure}

Fig.~\ref{fig:real_sim_half_gridrot} reports grid error (GE, left) and rotation error (RE, right) for sparse acquisitions with 12, 24, and 32 diffusion directions in the real-image recombination experiment. GraphSVR achieves significantly lower GE than \textit{eddy} across all sparsity levels. 

Importantly, the recombination protocol introduces abrupt stack-wise pose changes by drawing slice groups from different scans, thereby violating the assumption of smooth, continuous motion. The consistent improvements under this setting indicate that GraphSVR remains robust to non-smooth motion with rapid inter-stack changes.

\subsection{Fully synthetic data}

\begin{figure}[t]
    \centering
    \includegraphics[width=0.9\linewidth]{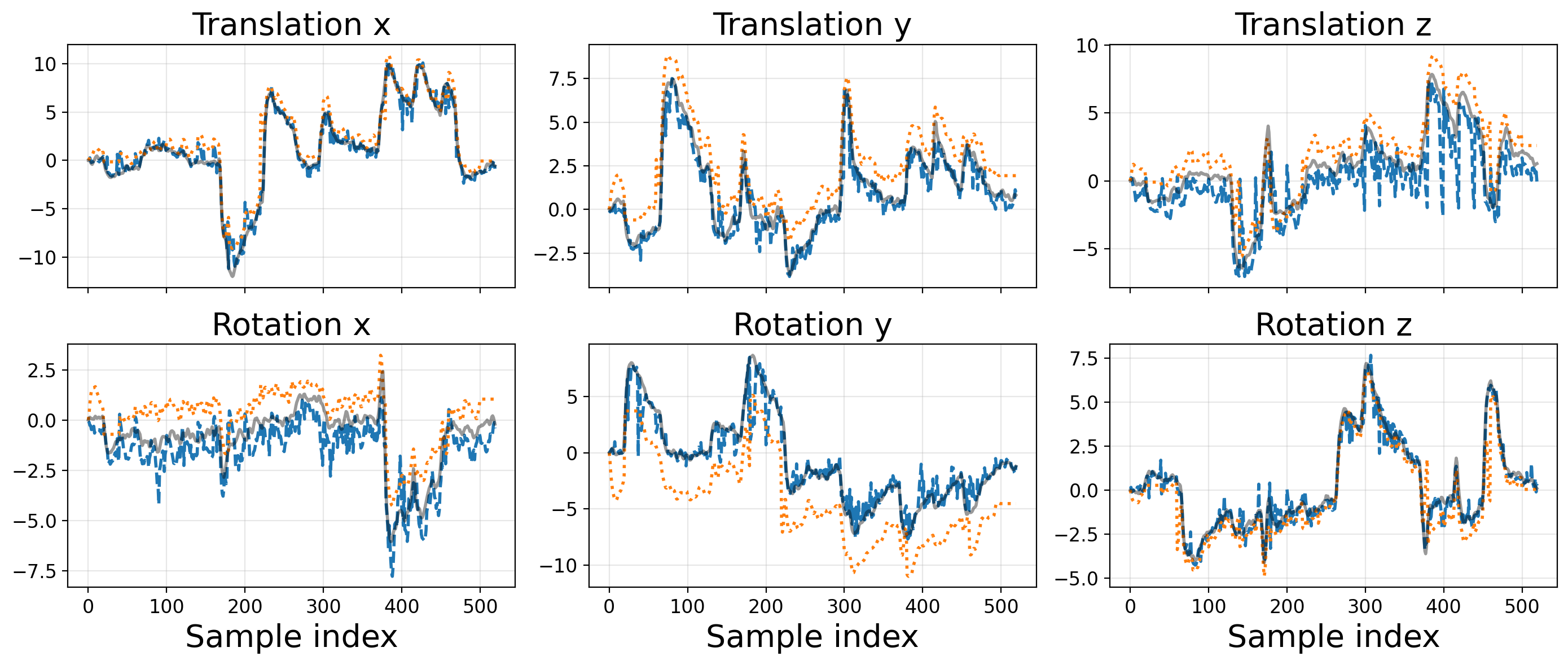}

    \caption{\textbf{Example rigid-motion trajectory (moderate synthetic motion).}
Ground-truth motion (black) compared with GraphSVR (blue) and \textit{eddy} (orange). Translations are shown in mm and rotations in degrees.}
    \label{fig:example_rigid_params}
\end{figure}

Tables~\ref{tab:full_sim_motion} and~\ref{tab:full_sim_ndirs} report mean$\pm$std of the per-case median errors as a function of motion severity and number of diffusion directions, respectively. Fig.~\ref{fig:example_rigid_params} illustrates a representative moderate-motion trajectory. GraphSVR follows the ground-truth motion more closely than \textit{eddy}, particularly during rapid pose changes, consistent with improved robustness.

Table~\ref{tab:full_sim_motion} shows that under severe motion, GraphSVR achieves significantly lower errors than \textit{eddy}. Moreover, GraphSVR maintains comparable error levels across motion severities, indicating stability under varying motion conditions. Notably, grid error remains below 3\,mm even in the severe setting.

\begin{table}[t]
\caption{\textbf{Fully synthetic simulation: motion severity.}
Mean$\pm$std over cases of per-case median errors. GE in mm, RE in radians. Bold indicates GraphSVR significantly outperforms \textit{eddy} ($p<0.05$, paired Wilcoxon test).
}
\label{tab:full_sim_motion}
\centering
\scriptsize
\setlength{\tabcolsep}{1.5pt}
\renewcommand{\arraystretch}{1.12}
\begin{tabular}{l cc cc cc}
\toprule
& \multicolumn{2}{c}{\textbf{Minor}}
& \multicolumn{2}{c}{\textbf{Moderate}}
& \multicolumn{2}{c}{\textbf{Severe}} \\
\cmidrule(lr){2-3}\cmidrule(lr){4-5}\cmidrule(lr){6-7}
\textbf{Method}
& \textbf{GE} & \textbf{RE}
& \textbf{GE} & \textbf{RE}
& \textbf{GE} & \textbf{RE} \\
\midrule
\makecell[l]{No\\ correction}
& $4.183\!\pm\!0.511$ & $0.042\!\pm\!0.009$
& $8.709\!\pm\!1.047$ &$0.091\!\pm\!0.015$
& $15.875\!\pm\!1.201$ & $0.163\!\pm\!0.021$ \\
\textit{eddy}
& $3.435\!\pm\!1.864$ & $0.035\!\pm\!0.024$
& $2.778\!\pm\!1.580$ & $0.023\!\pm\!0.020$
& $11.018\!\pm\!8.119$ & $0.119\!\pm\!0.087$ \\
\textbf{GraphSVR}\\ \textbf{(ours)}
& $2.084\!\pm\!1.846$ & $0.0221\!\pm\!0.0230$
& $2.277\!\pm\!2.257$ & $0.025\!\pm\!0.028$
& $\mathbf{2.952\!\pm\!3.395}$ & $\mathbf{0.031\!\pm\!0.037}$ \\
\bottomrule
\end{tabular}
\end{table}

\begin{table}[t]
\caption{\textbf{Fully synthetic simulation: gradient-direction count.}
Mean$\pm$std over cases of per-case median-over-stacks errors. GE in mm, RE in radians.}
\label{tab:full_sim_ndirs}
\centering
\scriptsize
\setlength{\tabcolsep}{1.5pt}
\renewcommand{\arraystretch}{1.12}
\begin{tabular}{l cc cc cc}
\toprule
& \multicolumn{2}{c}{\textbf{12 dirs}}
& \multicolumn{2}{c}{\textbf{24 dirs}}
& \multicolumn{2}{c}{\textbf{32 dirs}} \\
\cmidrule(lr){2-3}\cmidrule(lr){4-5}\cmidrule(lr){6-7}
\textbf{Method}
& \textbf{GE} & \textbf{RE}
& \textbf{GE} & \textbf{RE}
& \textbf{GE} & \textbf{RE} \\
\midrule
\makecell[l]{No\\ correction}
& $9.160\!\pm\!1.394$ & $0.098\!\pm\!0.013$
& $8.709\!\pm\!1.047$ & $0.091\!\pm\!0.015$
& $8.901\!\pm\!0.961$ & $0.088\!\pm\!0.011$ \\
\textit{eddy}
& $3.217\!\pm\!1.985$ & $0.034\!\pm\!0.021$
& $2.778\!\pm\!1.580$ & $0.030\!\pm\!0.020$
& $4.341\!\pm\!2.915$ & $0.045\!\pm\!0.035$ \\
\textbf{GraphSVR }\\ \textbf{(ours)}
& $2.217\!\pm\!2.067$ & $0.023\!\pm\!0.023$
& $2.188\!\pm\!2.057$ & $0.023\!\pm\!0.024$
& $2.417\!\pm\!2.556$ & $0.022\!\pm\!0.022$ \\
\bottomrule
\end{tabular}
\end{table}

% \subsection{Ablation study}
% We evaluate the contribution of edge features by selectively removing components used for graph construction and weighting. Results on four fully synthetic cases with moderate motion are reported in Table~\ref{tab:ablation_edge_features}. The full edge specification achieves the lowest GE and RE, demonstrating the complementary contribution of temporal, spatial, and q-space information.

\subsection{Ablation study}
We evaluate the contribution of individual edge features by selectively zeroing components of the joint acquisition space used for kNN graph construction and edge weighting. Experiments are conducted on 4 cases from the fully synthetic dataset (24 diffusion directions + two $b{=}0$ images) under moderate motion. Table~\ref{tab:ablation_edge_features} reports mean$\pm$std across cases of per-case median-over-stacks grid error (GE) and rotation error (RE). The full edge specification achieves the lowest errors, indicating that temporal, spatial, and q-space information provide complementary constraints. Removing all relational features substantially degrades performance, highlighting the importance of graph-based modeling.

\begin{table}[h!]
\caption{\textbf{Ablation of graph edge components.} Each row excludes selected features before edge-weight computation. Values show mean$\pm$std across cases of per-case median-over-stacks errors.}
\label{tab:ablation_edge_features}
\centering
\scriptsize
\setlength{\tabcolsep}{5pt}
\renewcommand{\arraystretch}{1.15}
\begin{tabular}{cccccc}
\toprule
\textbf{Q-space} & \textbf{Time} & \textbf{Location} &  \textbf{GE (mm)} & \textbf{RE (radians)} \\ 
\midrule
% \cmark & \cmark & \cmark & \cmark & $1.732 \pm 0.334$ & $0.017 \pm 0.002$ \\
\cmark & \cmark & \cmark &  $ 1.649 \pm 0.260$ & $0.016 \pm 0.001$ \\
\cmark & \cmark & \xmark & $1.676 \pm 0.268$ & $0.016 \pm 0.002$ \\
\cmark & \xmark & \xmark &  $1.728 \pm 0.297$ & $0.017 \pm 0.003$ \\
\xmark & \xmark & \xmark &  $ 3.202 \pm 2.417$ & $0.032 \pm 0.026$ \\
\bottomrule
\end{tabular}

\end{table}

\section{Conclusion}

We introduced \textbf{GraphSVR}, a q-space–aware graph-based framework for slice-to-volume motion estimation in DWI that explicitly models acquisition structure through a weighted graph over time, slice location, and diffusion encodings. By coupling diffusion measurements via graph message passing and optimizing in a self-supervised, zero-shot manner using only an anatomical reference, GraphSVR produces globally consistent stack-wise rigid motion estimates.

Across realistic recombination-based simulations and fully synthetic experiments with known ground truth, GraphSVR substantially outperforms FSL \textit{eddy}, particularly under severe motion, achieving large reductions in both grid and rotation errors. These findings demonstrate that explicitly modeling acquisition relationships through graph-based reasoning enables robust 4D DWI motion correction, especially in motion-prone settings.

\subsubsection*{Acknowledgements}
This work was supported in part by research grants from the Israel Innovation Authority (Kamin no. 90044/90045) and the United States-Israel Binational Science Foundation (BSF), Jerusalem, Israel (Award number 2025119). N.K is sponsored in part by a PhD scholarship for outstanding students in Data Sciences from the Israel Council for Higher Education.

\bibliographystyle{splncs04}
\bibliography{refs}

\end{document}